\documentclass{article} 
\usepackage{iclr2027_conference,times}

\usepackage{amsmath,amsfonts,bm}

\def\eqref#1{equation~\ref{#1}}

\def\1{\bm{1}}

\DeclareMathAlphabet{\mathsfit}{\encodingdefault}{\sfdefault}{m}{sl}
\SetMathAlphabet{\mathsfit}{bold}{\encodingdefault}{\sfdefault}{bx}{n}

\usepackage{hyperref}
\usepackage{url}
\usepackage[utf8]{inputenc} 
\usepackage[T1]{fontenc}    
\usepackage{hyperref}       
\usepackage{url}            
\usepackage{booktabs}       
\usepackage{amsfonts}       
\usepackage{nicefrac}       
\usepackage{microtype}      
\usepackage{xcolor}         

\usepackage[noend]{algpseudocode}
\usepackage{microtype}
\usepackage{graphicx}
\usepackage{subfigure}
\usepackage{color}

\usepackage{multirow}
\usepackage{float}
\floatstyle{plaintop}
\restylefloat{table}
\usepackage{caption}
\usepackage{enumitem}

\usepackage{algorithm}
\usepackage{algpseudocode}
\usepackage[dvipsnames, table]{xcolor}
\usepackage{caption}
\usepackage[tableposition=top]{caption}
\usepackage{pifont}
\usepackage{mathtools}

\usepackage{booktabs}
\usepackage{multirow}
\usepackage{wrapfig}

\usepackage{amsmath}
\usepackage{amssymb}

\usepackage{booktabs}
\usepackage{graphicx}
\usepackage{amssymb}

\usepackage{makecell}

\usepackage[textsize=tiny]{todonotes}
\usepackage{ifthen}
\newboolean{comments}
\setboolean{comments}{false} 

\title{Beyond Interaction Capacity: Estimator Scaling with Recursive Models for CTR Prediction}

\author{Shivang Chopra\thanks{Work done during internship at Google Research} \\
Georgia Institute of Technology \\
\texttt{shivangchopra11@gatech.edu} \\
\And
Fotis Iliopoulos \\
Google Research \\
\texttt{fotisi@google.com} \\
\And
Zsolt Kira \\
Georgia Institute of Technology \\
\texttt{zkira@gatech.edu} \\
\And
Gaurav Menghani \\
Google Research \\
\texttt{gmenghani@google.com} 
}

\iclrfinalcopy 
\begin{document}

\maketitle

\begin{abstract}
Click-Through Rate (CTR) prediction, a core task in recommendation and advertising systems, relies on modeling interactions among sparse categorical features. Explicit cross networks are a central paradigm for CTR prediction, and recent progress has largely come from increasing the interaction capacity of a single predictor through deeper cross networks and more expressive cross operators. We revisit whether continually increasing interaction capacity remains the most effective way to improve predictive performance, and find that its benefits quickly exhibit diminishing returns even as capacity continues to grow. This motivates a complementary scaling direction that we call \emph{estimator scaling}, where additional resources are used to incorporate multiple related estimators rather than only enlarging a single predictor. Through theoretical analysis and diagnostic experiments, we show that the gains from estimator scaling are governed by the amount of non-shared predictive variation available across estimators. However, exploiting this variation naively can be expensive: independently trained models provide substantial estimator diversity but require deployment cost to grow with ensemble size. This motivates a parameter-efficient realization of estimator scaling that can incorporate diversity from multiple estimator sources without maintaining multiple full models. Building on this view, we introduce \textbf{REC}ursive \textbf{A}veraged \textbf{P}redictor (RECAP), a parameter-efficient recursive CTR model that operationalizes estimator scaling at three levels: distillation across independently trained models, exponential moving averaging over training trajectories, and aggregation over inference-time routes within a weight-shared recursive backbone. This design improves predictive performance without requiring parameter growth proportional to the ensemble size. Experiments across multiple CTR benchmarks show that RECAP approaches the accuracy of a five-model ensemble of a strong cross-network baseline at one fifth of its deployed parameters, while doubling the baseline's interaction capacity recovers only a third of that gain. Overall, these results establish new state-of-the-art predictive performance on standard  benchmarks, while placing the proposed approach on a favorable performance–parameter Pareto frontier. 

\end{abstract}
\vspace{-15pt}
\section{Introduction}
\label{sec:intro}
\vspace{-5pt}

Click-through rate (CTR) prediction is a central component of modern recommendation and advertising systems, where accurate ranking depends critically on modeling interactions among sparse categorical and numerical features~\cite{10.1145/3477495.3531723,10.1145/3124749.3124754}. A long line of work has therefore improved CTR models by increasing the \emph{interaction capacity} of a single predictor through deeper cross networks, and richer interaction operators~\cite{10.1145/3124749.3124754,10.1145/3442381.3450078,3172077.3172127,10.1145/3539618.3591988}. In explicit cross networks, this capacity can be increased along several architectural dimensions, most directly by making interaction networks deeper or by combining multiple interaction branches with complementary operators. Deep \& Cross Networks (DCNs) \cite{10.1145/3124749.3124754} and their derivatives exemplify this paradigm, explicitly scaling the complexity of feature interactions represented by the model. Recent architectures such as Quadratic Neural Networks (QNN) and Fusing Cross Networks (FCN) push this direction further by constructing interactions whose order grows exponentially with depth~\cite{10.1145/3711896.3737106,10.1145/3770854.3780177}. This progression raises a natural question: \emph{once a sufficiently expressive interaction model is available, is continually increasing interaction capacity still the most effective way to improve CTR prediction?}

\begin{figure}[t]
    \centering
    \includegraphics[width=0.9\linewidth]{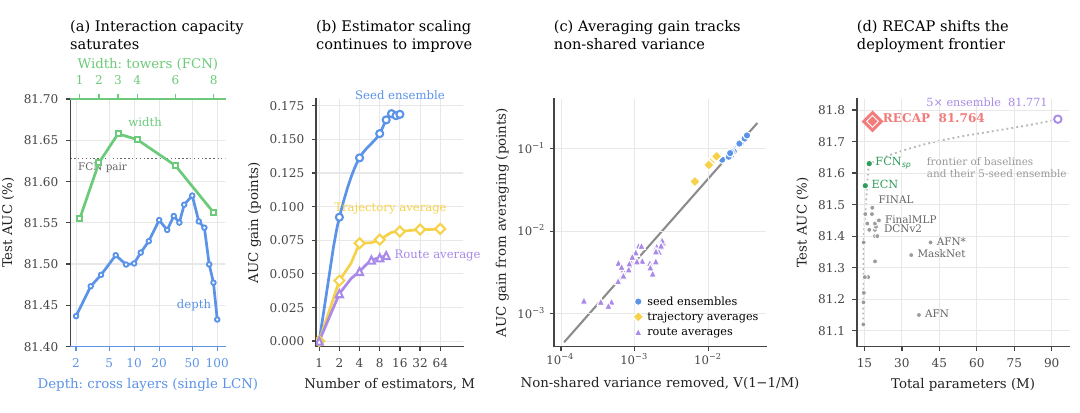}
    \vspace{-15pt}
    \caption{
    \textbf{Diagnostic analysis of interaction-capacity and estimator scaling on Criteo.}
    (a) Increasing interaction depth and breadth exhibits diminishing returns.
    (b) Additional estimators provide further gains, with independent runs outperforming trajectories and shared-model routes.
    (c) Averaging gains track the non-shared predictive variance available to aggregation (Section~\ref{sec:variance_scaling}).
    (d) RECAP captures much of the ensemble gain at approximately a single-model deployment budget. Diagnosis results on additional datasets can be found in Appendix \ref{app:diagnosis}.
    \vspace{-20pt}
    }
    \label{fig:intro}
\end{figure} 
We test this assumption by scaling interaction capacity along two concrete architectural axes: depth, through additional cross layers, and breadth, through additional interaction branches. As shown in Figure~\ref{fig:intro}(a), increasing interaction depth or adding interaction branches initially improves performance, but the marginal benefit quickly shrinks as interaction capacity increases. This diminishing-return regime raises a second question: if additional resources are no longer best spent increasing the capacity of a single predictor, how else can they improve prediction? 

In Section~\ref{sec:estimator_scaling}, we conduct a theoretical analysis that points to a complementary scaling direction that we call \textbf{estimator scaling}. Rather than enlarging the function class of a single predictor, estimator scaling uses additional training or inference resources to incorporate multiple estimators of the same predictive function. Prior work has theoretically analyzed the underlying benefit of averaging less-correlated estimators \cite{JMLR:v24:23-0041}; our focus is on its role as a scaling principle for CTR models, where recent progress has predominantly come from increasing feature-interaction capacity. We show that its benefit is governed by the non-shared predictive variation exposed by different estimator sources which offer distinct diversity--deployment-cost trade-offs. This yields a simple scaling principle: gains are larger when additional estimators expose more non-shared variation and diminish as their predictions become increasingly correlated.

As shown in Figure~\ref{fig:intro}(b), our diagnostic experiments support this prediction across three distinct sources of estimators: averaging independently trained models, averaging predictors sampled along a single training trajectory, and averaging alternative inference-time routes through a shared model. Despite their different origins, the improvement from averaging closely tracks the amount of \emph{non-shared predictive variation} available along each estimator axis (Figure~\ref{fig:intro}(c)). Additional diagnosis results on other datasets can be found in Appendix \ref{app:diagnosis}. These results reveal both the promise and the challenge of estimator scaling: large gains arise from averaging multiple estimators but maintaining individual copies of the models would incur substantial deployment cost.

To address these challenges, we introduce \textbf{RECursive Averaged Predictor (RECAP)}, a parameter-efficient recursive CTR model that realizes estimator scaling at three levels. RECAP uses a weight-shared recursive interaction backbone, allowing interaction computation to increase without proportional growth in depth-specific parameters. On top of this backbone, RECAP incorporates information across \emph{independent training runs} through ensemble distillation, across the \emph{training trajectory} through exponential moving weight averaging, and across \emph{inference-time routes} through route aggregation.  As summarized in Figure~\ref{fig:intro}, these mechanisms expose progressively different sources of estimator variation while keeping the deployed model close to a single-model parameter budget, thereby placing RECAP on a favorable deployment frontier.

Our main contributions are as follows:
\begin{itemize}
    \vspace{-10pt}
    \item \textbf{Estimator scaling for CTR prediction.}
    We identify estimator scaling as a complementary axis to interaction-capacity scaling and show that estimator scaling continues to provide gains after additional interaction depth and branch capacity exhibit diminishing returns.

    \item \textbf{RECAP: parameter-efficient multi-level estimator scaling.}
    We introduce a recursive CTR model that combines parameter sharing with estimator scaling across training trajectories, independent runs, and inference-time routes through exponential moving weight averaging, ensemble distillation, and route aggregation.

    \item \textbf{Improved deployment efficiency.}
    Across multiple CTR benchmarks, RECAP improves the performance--parameter trade-off over strong cross-network baselines and approaches the performance of substantially larger multi-model ensembles while retaining approximately a single-model deployment footprint.

\end{itemize}

\section{Related Work}

\textbf{Feature Interaction and Cross-Network Scaling: }A central line of CTR research improves predictive accuracy by scaling the learned feature interactions. Early models combine factorization-based interactions with deep networks~\cite{3172077.3172127}, while DCN explicitly constructs bounded-degree feature crosses through stacked cross layers~\cite{10.1145/3124749.3124754}. Subsequent work increases or adapts this interaction capacity in different ways: xDeepFM introduces explicit vector-wise high-order interactions~\cite{10.1145/3219819.3220023}, AutoInt uses stacked self-attention to model increasingly high-order combinations~\cite{10.1145/3357384.3357925}, and DCNv2 improves the expressiveness of cross networks while retaining computational efficiency through low-rank parameterizations~\cite{10.1145/3442381.3450078}. Most recently, FCN and QNN use exponential cross networks to explicitly represent interactions spanning a broad range of orders~\cite{10.1145/3711896.3737106,10.1145/3770854.3780177}. Collectively, these methods primarily improve CTR prediction by enlarging or selectively allocating the \emph{interaction capacity} of a predictor. In contrast, we study what should scale once further increases along this axis yield diminishing predictive returns.

\textbf{Ensembling, Averaging, and Distillation: }Our work is also related to methods that combine multiple learned predictors. Deep ensembles improve prediction by aggregating independently trained models~\cite{3295222.3295387}, while trajectory-based methods such as stochastic weight averaging (SWA) aggregate solutions encountered during optimization~\cite{izmailov2019averagingweightsleadswider}. Model Soups and DiWA further study weight averaging across independently trained or fine-tuned models, highlighting the importance of model diversity for successful aggregation~\cite{pmlr-v162-wortsman22a,rame2022diwa}. The role of diversity in ensemble performance has also been studied theoretically~\cite{JMLR:v24:23-0041}. Our contribution is not a new averaging operator in isolation. Instead, we formulate \emph{estimator scaling} as a complementary scaling axis and study its gains, saturation, and cost across independent runs, optimization trajectories, and inference-time routes within a common variance-based framework

\textbf{Recursive and Parameter-Efficient Computation: }A complementary literature seeks to increase computation or interaction complexity without proportional parameter growth. Low-rank cross networks reduce the parameter cost of expressive interaction operators~\cite{10.1145/3442381.3450078}, while recent LoopCTR introduces loop scaling for CTR prediction, recursively reusing shared layers to decouple additional computation from model size~\cite{tang2026loopctrunlockingloopscaling}. RECAP shares the principle of recursive parameter reuse but serves a different purpose: recursion provides a compact substrate on which multiple related estimators can be realized. We combine this backbone with cross-run distillation, trajectory averaging, and inference-time route aggregation, using parameter sharing to enable parameter-efficient estimator scaling.

\vspace{-5pt}
\section{From Interaction Capacity to Estimator Scaling}
\label{sec:estimator_scaling}
\vspace{-5pt}
In this section, we formalize why estimator scaling can provide a complementary source of improvement to increasing the interaction capacity of a single predictor, characterize when such gains are available, and analyze how they combine across different estimator sources.

Consider a CTR dataset $\mathcal{D}=\{(x_i,y_i)\}_{i=1}^{N}$, where $x_i$ contains categorical and numerical features and $y_i\in\{0,1\}$ denotes a click. A model with parameters $\theta$ produces a logit $z$ and click probability $p_\theta(x)=\sigma(z)$, and is trained using binary cross-entropy, which is defined as:

\begin{equation}
    \ell(y,z)=\log(1+\exp(z))-yz.
    \label{eq:bce_logit}
\end{equation}

A fixed model family can yield multiple related predictors through independent training runs, checkpoints along an optimization trajectory, or alternative inference-time routes. We refer to each such predictor as an \emph{estimator}, and denote these three estimator sources by $S$, $T$, and $R$, respectively. Given $M$ estimators with logits $\{Z_m(x)\}_{m=1}^{M}$, their mean-logit prediction is
\begin{equation}
    \bar Z_M(x)=\frac{1}{M}\sum_{m=1}^{M} Z_m(x).
    \label{eq:estimator_scaling}
\end{equation}

\paragraph{Two Complementary Scaling Axes}
\label{sec:scaling_complementarity}

As motivated in the introduction, \textbf{interaction-capacity scaling} increases the expressive capacity of a single predictor, whereas \textbf{estimator scaling} increases the number of estimators incorporated into its prediction while largely holding the underlying model family fixed. We next formalize why these two axes can provide distinct sources of improvement.

Let $c$ index the interaction capacity of a model family, let $Z_c(x)$ denote the logit of a randomly drawn estimator from that family, and let $\mu_c(x)=\mathbb{E}[Z_c(x)\mid x]$ denote the estimator-family mean. For $M$ estimators, define $\bar Z_{c,M}(x)=\frac{1}{M}\sum_{m=1}^{M}Z_{c,m}(x)$, and let $\mathcal{L}(c,M)$ denote its expected binary cross-entropy over the data distribution and estimator randomness. We similarly define $\mathcal{L}(\mu_c)=\mathbb{E}_{x,y}[\ell(y,\mu_c(x))]$. A second-order expansion of Eq.~\ref{eq:bce_logit} around $\mu_c(x)$ gives
\begin{equation}
    \mathcal{L}(c,M)
    \approx
    \mathcal{L}(\mu_c)
    +
    \frac{1}{2}
    \mathbb{E}_{x}
    \left[
        \sigma(\mu_c(x))
        \left(1-\sigma(\mu_c(x))\right)
        \operatorname{Var}(\bar Z_{c,M}(x)\mid x)
    \right].
    \label{eq:capacity_estimator_decomposition}
\end{equation}

Equation~\ref{eq:capacity_estimator_decomposition} separates two mechanisms for improvement. Increasing interaction capacity can improve the mean predictor $\mu_c$ and thereby reduce $\mathcal{L}(\mu_c)$, whereas estimator scaling reduces variation around that mean through aggregation. Thus, even when additional interaction capacity yields little further improvement in the mean predictor, estimator scaling can still reduce expected loss whenever reducible estimator variation remains.

\paragraph{Non-Shared Variation Governs Estimator Scaling}
\label{sec:variance_scaling}

We next characterize how much estimator variation can be removed by aggregation. For a fixed input $x$ and interaction capacity $c$, suppose the $M$ estimators are exchangeable (i.e their joint probability distribution is invariant under any permutation of their indices) , with conditional logit variance $\tau_c^2(x)$ and common pairwise correlation $\rho_c(x)$. The variance of their mean is
\begin{equation}
    \operatorname{Var}(\bar Z_{c,M}(x)\mid x)
    =
    \tau_c^2(x)
    \left(
        \rho_c(x)+\frac{1-\rho_c(x)}{M}
    \right).
    \label{eq:equicorr_variance}
\end{equation}

Relative to a single estimator, the variance removed by averaging is therefore
\begin{equation}
    V_{\mathrm{ns}}(c,M;x)
    =
    \tau_c^2(x)
    \left(1-\rho_c(x)\right)
    \left(1-\frac{1}{M}\right),
    \label{eq:variance_removed}
\end{equation}
which we call the \emph{non-shared}, or reducible, predictive variation. Combining Eqs.~\ref{eq:capacity_estimator_decomposition}
and~\ref{eq:variance_removed}, the reduction in expected BCE loss is locally
\begin{equation}
    \Delta\mathcal{L}(c,M)
    \approx
    \frac{1}{2}
    \mathbb{E}_{x}
    \left[
        \sigma(\mu_c(x))
        \left(1-\sigma(\mu_c(x))\right)
        V_{\mathrm{ns}}(c,M;x)
    \right].
    \label{eq:curvature_weighted_variance}
\end{equation}
Thus, BCE improvement is approximately proportional to the reducible
predictive variance, up to the local curvature of the loss. When the
estimator families have similar mean predictions, this curvature factor is
approximately shared, so differences in averaging gain are primarily
determined by $V_{\mathrm{ns}}$. Further derivations of the relationship between non-shared variance and estimator scaling are provided in Appendix \ref{app:variance_theory}.

\paragraph{Combining Estimator Axes}
\label{sec:residual_variance}

Different estimator sources may expose overlapping predictive variation, so their gains need not be additive. Let $A$ and $B$ denote two estimator axes and let $Z_{A,B}(x)$ denote the corresponding random logit for a fixed input. By the law of total variance,
\begin{equation}
    \operatorname{Var}(Z_{A,B}\mid x)
    =
    \underbrace{
        \mathbb{E}_{B}
        \left[
            \operatorname{Var}_{A}(Z_{A,B}\mid B,x)
        \right]
    }_{V_A(x)}
    +
    \underbrace{
        \operatorname{Var}_{B}
        \left[
            \mathbb{E}_{A}(Z_{A,B}\mid B,x)
        \right]
    }_{V_{B\mid A}(x)}.
    \label{eq:two_axis_total_variance}
\end{equation}

Here $V_A(x)$ is the variation accessible to averaging along axis $A$, whereas $V_{B\mid A}(x)$ is the residual variation remaining after averaging over $A$. Up to the same local curvature weighting as in Eq.~\ref{eq:curvature_weighted_variance}, the standalone reduction in expected loss from an estimator axis is therefore governed by its accessible variation, while the marginal improvement from adding another axis is governed by the residual variation not already removed. This prediction is evaluated empirically in Section~\ref{sec:variance_results}; the full three-axis decomposition for independent runs, trajectories, and routes is given in Appendix~\ref{app:multi_axis_variance}

\section{RECAP: RECursive Averaged Predictor}
\label{sec:recap}
\vspace{-10pt}

We now introduce \textbf{REC}ursive \textbf{A}veraged \textbf{P}redictor (RECAP), a parameter-efficient realization of the estimator-scaling principle in Section~\ref{sec:estimator_scaling}. The theory identifies non-shared and residual predictive variation as the quantities governing the benefit of additional estimator axes; RECAP focuses on realizing three such axes: independent runs, optimization trajectories, and inference-time routes, without deploying a separate full model for each. These axes connect to the output-space analysis in different ways: ensemble distillation compresses a cross-run mean-logit predictor into a single model, EMA provides a parameter-space approximation to trajectory prediction averaging, and route aggregation performs the mean-logit averaging analyzed directly in Section~\ref{sec:estimator_scaling}.

\begin{figure*}[t]
    \centering
    \includegraphics[width=0.95\linewidth]{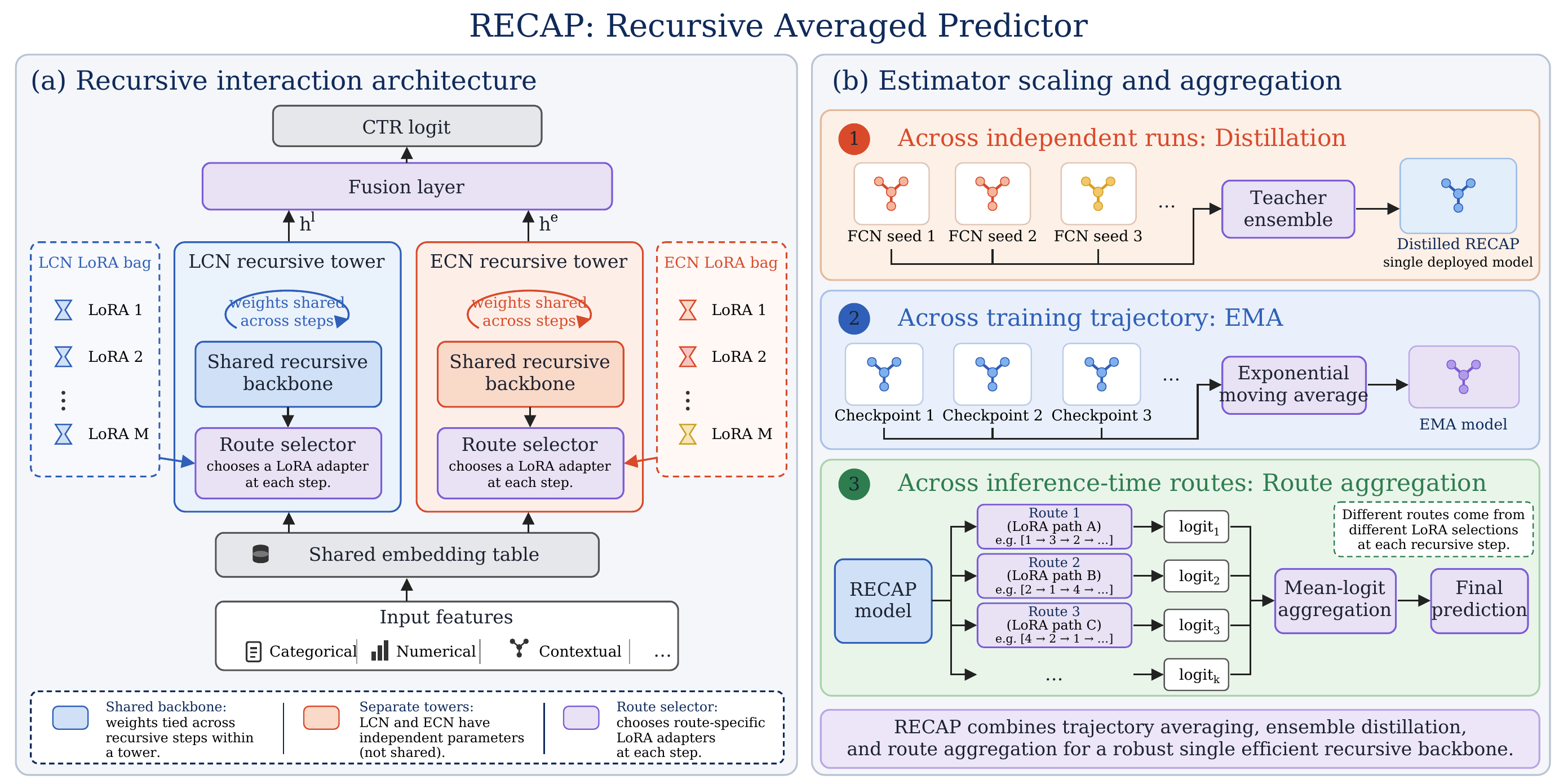}
    \vspace{-10pt}
    \caption{
\textbf{RECAP architecture and multi-level estimator scaling.}
(a) RECAP builds on the linear and exponential cross-network (LCN and ECN) operators of FCN~\citep{10.1145/3770854.3780177}, which we organize as two recursive interaction towers. Each tower shares its core cross-network parameters across recursive steps, while a route selector chooses lightweight LoRA adaptations from a tower-specific expert bank. The two tower representations are fused to produce the CTR logit.
(b) RECAP incorporates estimators at three levels: cross-run information is transferred through ensemble distillation, within-run information is aggregated through exponential moving averaging, and multiple recursive routes are combined through mean-logit aggregation at inference.
}
    \vspace{-15pt}
    \label{fig:recap_arch}
\end{figure*}

\vspace{-10pt}
\subsection{Parameter-Efficient Recursive Interaction Backbone}
\label{sec:recap_backbone}
\vspace{-8pt}

The estimator-scaling analysis does not prescribe a particular CTR backbone; it specifies that useful estimator axes should expose non-shared predictive variation at acceptable cost. We therefore use recursion for two complementary purposes: to increase interaction computation through parameter sharing, and to expose an additional inference-time estimator axis through alternative computation routes. We instantiate RECAP using the Fusing Cross Network (FCN) architecture~\citep{10.1145/3770854.3780177}, a strong recent explicit-interaction model for CTR prediction. FCN combines a linear cross network (LCN), whose interaction order grows progressively with depth, with an exponential cross network (ECN), which constructs higher-order interactions more rapidly. Given input features $x$, RECAP first maps sparse categorical fields and numerical features to a shared representation $h_0$, which is then processed by recursive LCN and ECN towers. The two towers use independent parameters, while the core interaction weights within each tower are shared across recursive steps. Lightweight route-specific adaptations allow different adapter sequences to define multiple related predictors over the same shared backbone, providing a third source of estimator variation that can be aggregated at inference without maintaining separate full models. Thus, recursion serves both as a parameter-efficient computational substrate and as the mechanism that enables RECAP's route-scaling axis. Additionally, the broader estimator-scaling principle is not tied to these particular interaction operators; Section~\ref{sec:fcn_scaling} further examines estimator scaling independently of the recursive architecture.

Let $q\in\{\mathrm{L},\mathrm{E}\}$ denote a tower and let $h_t^q$ be its hidden representation after $t$ recursive steps. RECAP updates
\begin{equation}
    h_{t+1}^{q}
    =
    F_q
    \left(
        h_t^{q},x;
        W_q+\Delta W_{q,r_t}
    \right),
    \label{eq:recap_recursive_update}
\end{equation}
where $W_q$ is the shared interaction operator and $\Delta W_{q,r_t}$ is a lightweight route-dependent adaptation selected at step $t$.

Rather than assigning a full independent parameter matrix to every recursive step, RECAP parameterizes each adaptation using a low-rank residual,
\begin{equation}
    \Delta W_{q,r}
    =
    A_{q,r}B_{q,r}^{\top},
    \qquad
    \operatorname{rank}(\Delta W_{q,r})\leq r_{\mathrm{LoRA}}.
    \label{eq:recap_lora}
\end{equation}
Each tower maintains a small bank of such residuals, and a route selector determines which adapter is applied at each recursive step. A route is therefore defined by the sequence $\tau_q = (r_1,r_2,\ldots,r_T),$ while all routes continue to reuse the same core operator $W_q$.

After $T$ recursive steps, the two tower representations are fused, $h_{\mathrm{fuse}} =
    \operatorname{Fuse}
    \left(
        h_T^{\mathrm{L}},
        h_T^{\mathrm{E}}
    \right),$ and a prediction head produces the CTR logit $ z_\theta(x) = g(h_{\mathrm{fuse}})$. This recursive parameterization decouples interaction computation from parameter depth: increasing the number of recursive steps increases computation while reusing the same backbone parameters. The lightweight route residuals additionally allow RECAP to expose multiple related predictors without storing multiple models.

\subsection{Estimator Scaling at Three Levels}
\label{sec:recap_three_levels}

RECAP applies the estimator-scaling principle of Section~\ref{sec:estimator_scaling} at three complementary levels.

\textbf{Training-trajectory averaging: }During optimization, successive checkpoints provide different estimators from the same training run. RECAP maintains an exponential moving average (EMA) of the model parameters,
\begin{equation}
    \bar\theta_t
    =
    \beta\bar\theta_{t-1}
    +
    (1-\beta)\theta_t,
    \label{eq:recap_ema}
\end{equation}
where $\beta$ controls the averaging horizon. The EMA parameters are used for validation, checkpoint selection, and final inference. This incorporates information across the optimization trajectory without adding deployed parameters or inference-time computation.

EMA averages parameters rather than logits directly. Its connection to the output-space analysis of Section~\ref{sec:estimator_scaling} follows under a local linearization of the predictor. If
$\bar{\theta}=\sum_t \alpha_t\theta_t$ denotes the EMA parameters, then for checkpoints lying in a locally approximately linear region,
\begin{equation}
z_{\bar{\theta}}(x)
\approx
\sum_t \alpha_t z_{\theta_t}(x).
\end{equation}
Thus, EMA can be viewed as a parameter-efficient approximation to trajectory-level prediction averaging. This approximation need not hold globally, so we empirically evaluate trajectory scaling rather than treating it as an exact consequence of the logit-averaging analysis.

\textbf{Cross-run averaging through distillation:} Independent training runs expose substantially more non-shared predictive variation, but deploying an ensemble of $M$ complete CTR models multiplies storage and inference cost. RECAP instead transfers this cross-run aggregation into a single model through ensemble distillation.

Let $z^{(1)}(x),\ldots,z^{(M)}(x)$ denote the logits of independently trained teacher models. We form the teacher aggregate
\begin{equation}
    \bar z_{\mathrm{teach}}(x)
    =
    \frac{1}{M}
    \sum_{m=1}^{M}
    z^{(m)}(x),
    \label{eq:teacher_mean}
\end{equation}
and train RECAP to match the corresponding soft prediction in addition to the ground-truth label. The training objective is
\begin{equation}
    \mathcal{L}_{\mathrm{train}}
    =
    \mathcal{L}_{\mathrm{sup}}
    +
    \lambda_{\mathrm{KD}}
    \mathcal{L}_{\mathrm{KD}},
    \label{eq:recap_training_loss}
\end{equation}
where
\begin{equation}
    \mathcal{L}_{\mathrm{KD}}
    =
    \operatorname{BCE}
    \left(
        \sigma(\bar z_{\mathrm{teach}}),
        z_\theta(x)
    \right).
\end{equation}
Thus, distillation transfers the benefit of expensive cross-run averaging into a single deployed parameterization.

This objective has a direct connection to the mean-logit predictor analyzed in Section~\ref{sec:estimator_scaling}. For a fixed input, BCE with soft target $q=\sigma(\bar z_{\mathrm{teach}})$ is minimized when $\sigma(z_\theta)=q$, or equivalently $z_\theta=\bar z_{\mathrm{teach}}$. Thus, with sufficient student capacity and optimization, distillation compresses the cross-run mean-logit aggregate into a single deployed predictor. In practice this equality is approximate,
and the remaining compression error is measured empirically.

\textbf{Inference-time route aggregation: }Finally, the recursive architecture provides a third estimator source at inference. Different LoRA selections define different recursive routes $\tau_1,\ldots,\tau_K,$ which produce logits $z_{\tau_1}(x),\ldots,z_{\tau_K}(x).$

RECAP combines them using mean-logit aggregation,
\begin{equation}
    z_{\mathrm{RECAP}}^{(K)}(x)
    =
    \frac{1}{K}
    \sum_{k=1}^{K}
    z_{\tau_k}(x).
    \label{eq:route_aggregation}
\end{equation}
Increasing $K$ therefore increases inference compute but leaves the stored model unchanged. This provides a test-time scaling knob that is complementary to both EMA and cross-run distillation.
\vspace{-5pt}

\subsection{Training Across Recursive Budgets}
\label{sec:recap_training}
\vspace{-5pt}

Because the recursive backbone may be evaluated at different inference depths, we train RECAP across multiple rollout budgets rather than specializing it to a single depth. Let $\mathcal{D}_{\mathrm{train}}$ denote the set of training depths. The supervised objective is
\begin{equation}
    \mathcal{L}_{\mathrm{sup}}
    =
    \sum_{d\in\mathcal{D}_{\mathrm{train}}}
    \alpha_d\,
    \ell
    \left(
        y,
        z_\theta^{(d)}(x)
    \right),
    \label{eq:multibudget_loss}
\end{equation}
where $\alpha_d$ controls the contribution of rollout depth $d$. During training, routes are sampled from the LoRA banks so that the shared backbone is exposed to multiple recursive computation paths.

We optimize RECAP with a decaying learning-rate schedule and maintain the EMA parameters from Eq.~\ref{eq:recap_ema} throughout training. Teacher predictions are precomputed, so ensemble distillation does not require executing the teacher models during student optimization.

\vspace{-8pt}
\section{Experiments and Results}
\vspace{-8pt}

\begin{table*}[t]
\renewcommand\arraystretch{1.2}
\centering
\caption{\textbf{Performance comparison of different deep CTR models.} Typically, CTR researchers consider an improvement of \textit{0.001 (0.1\%)} in Logloss and AUC to be practically meaningful \cite{10.1145/3459637.3482486,10.1145/3770854.3780177}. We conduct a two-tailed t-test over five independent runs to assess the statistical significance between our models and the best baseline (*: p<0.01). Best results are highlighted in \textbf{bold}, second-best are \underline{underlined}. \vspace{-10pt}} 
\label{tab:main_results}

\resizebox{\linewidth}{!}{
\begin{tabular}{ccccccccccc}
\Xhline{1px}

\multicolumn{1}{c|}{} 
& \multicolumn{2}{c|}{\textbf{Avazu}} 
& \multicolumn{2}{c|}{\textbf{Criteo}} 
& \multicolumn{2}{c|}{\textbf{ML-1M}} 
& \multicolumn{2}{c|}{\textbf{KDD12}} 
& \multicolumn{2}{c}{\textbf{KKBox}} \\ 
\cline{2-11}

\multicolumn{1}{c|}{\multirow{-2}{*}{\textbf{Models}}} 
& \multicolumn{1}{c}{Logloss$\downarrow$} 
& \multicolumn{1}{c|}{AUC(\%)$\uparrow$} 
& \multicolumn{1}{c}{Logloss$\downarrow$} 
& \multicolumn{1}{c|}{AUC(\%)$\uparrow$} 
& \multicolumn{1}{c}{Logloss$\downarrow$} 
& \multicolumn{1}{c|}{AUC(\%)$\uparrow$} 
& \multicolumn{1}{c}{Logloss$\downarrow$} 
& \multicolumn{1}{c|}{AUC(\%)$\uparrow$} 
& \multicolumn{1}{c}{Logloss$\downarrow$} 
& \multicolumn{1}{c}{AUC(\%)$\uparrow$} \\

\hline

\multicolumn{1}{c|}{DNN \cite{10.1145/2959100.2959190}} 
& 0.3721 & \multicolumn{1}{c|}{79.27} 
& 0.4380 & \multicolumn{1}{c|}{81.40} 
& 0.3100 & \multicolumn{1}{c|}{90.30} 
& 0.1502 & \multicolumn{1}{c|}{80.52} 
& 0.4811 & \multicolumn{1}{c}{85.01} \\

\multicolumn{1}{c|}{PNN \cite{7837964}} 
& 0.3712 & \multicolumn{1}{c|}{79.44} 
& 0.4378 & \multicolumn{1}{c|}{81.42} 
& 0.3070 & \multicolumn{1}{c|}{90.42} 
& 0.1504 & \multicolumn{1}{c|}{80.47} 
& 0.4793 & \multicolumn{1}{c}{85.15} \\

\multicolumn{1}{c|}{Wide \& Deep \cite{10.1145/2988450.2988454}} 
& 0.3720 & \multicolumn{1}{c|}{79.29} 
& 0.4376 & \multicolumn{1}{c|}{81.42} 
& 0.3056 & \multicolumn{1}{c|}{90.45} 
& 0.1504 & \multicolumn{1}{c|}{80.48} 
& 0.4852 & \multicolumn{1}{c}{85.04} \\

\multicolumn{1}{c|}{DeepFM \cite{3172077.3172127}} 
& 0.3719 & \multicolumn{1}{c|}{79.30} 
& 0.4375 & \multicolumn{1}{c|}{81.43} 
& 0.3073 & \multicolumn{1}{c|}{90.51} 
& 0.1501 & \multicolumn{1}{c|}{80.60} 
& 0.4785 & \multicolumn{1}{c}{85.31} \\

\multicolumn{1}{c|}{DCNv1 \cite{10.1145/3124749.3124754}} 
& 0.3719 & \multicolumn{1}{c|}{79.31} 
& 0.4376 & \multicolumn{1}{c|}{81.44} 
& 0.3156 & \multicolumn{1}{c|}{90.38} 
& 0.1501 & \multicolumn{1}{c|}{80.59} 
& 0.4766 & \multicolumn{1}{c}{85.31} \\

\multicolumn{1}{c|}{xDeepFM \cite{10.1145/3219819.3220023}} 
& 0.3718 & \multicolumn{1}{c|}{79.33} 
& 0.4376 & \multicolumn{1}{c|}{81.43} 
& 0.3054 & \multicolumn{1}{c|}{90.47} 
& 0.1501 & \multicolumn{1}{c|}{80.62} 
& 0.4772 & \multicolumn{1}{c}{85.35} \\

\multicolumn{1}{c|}{AutoInt \cite{10.1145/3357384.3357925}} 
& 0.3746 & \multicolumn{1}{c|}{79.02} 
& 0.4390 & \multicolumn{1}{c|}{81.32} 
& 0.3112 & \multicolumn{1}{c|}{90.45} 
& 0.1502 & \multicolumn{1}{c|}{80.57} 
& 0.4773 & \multicolumn{1}{c}{85.34} \\

\multicolumn{1}{c|}{AFN \cite{cheng2020adaptivefactorizationnetworklearning}} 
& 0.3726 & \multicolumn{1}{c|}{79.29} 
& 0.4384 & \multicolumn{1}{c|}{81.38} 
& 0.3048 & \multicolumn{1}{c|}{90.53} 
& 0.1499 & \multicolumn{1}{c|}{80.70} 
& 0.4842 & \multicolumn{1}{c}{84.89} \\

\multicolumn{1}{c|}{DCNv2 \cite{10.1145/3442381.3450078}} 
& 0.3718 & \multicolumn{1}{c|}{79.31} 
& 0.4376 & \multicolumn{1}{c|}{81.45} 
& 0.3098 & \multicolumn{1}{c|}{90.56}
& 0.1502 & \multicolumn{1}{c|}{80.59} 
& 0.4787 & \multicolumn{1}{c}{85.31} \\

\multicolumn{1}{c|}{EDCN \cite{10.1145/3459637.3481915}} 
& 0.3716 & \multicolumn{1}{c|}{79.35} 
& 0.4386 & \multicolumn{1}{c|}{81.36} 
& 0.3073 & \multicolumn{1}{c|}{90.48} 
& 0.1501 & \multicolumn{1}{c|}{80.62} 
& 0.4952 & \multicolumn{1}{c}{85.27} \\

\multicolumn{1}{c|}{MaskNet \cite{wang2021masknetintroducingfeaturewisemultiplication}} 
& 0.3711 & \multicolumn{1}{c|}{79.43} 
& 0.4387 & \multicolumn{1}{c|}{81.34} 
& 0.3080 & \multicolumn{1}{c|}{90.34} 
& 0.1498 & \multicolumn{1}{c|}{80.79} 
& 0.5003 & \multicolumn{1}{c}{84.79} \\

\multicolumn{1}{c|}{EulerNet \cite{10.1145/3539618.3591681}} 
& 0.3723 & \multicolumn{1}{c|}{79.22} 
& 0.4379 & \multicolumn{1}{c|}{81.47} 
& 0.3050 & \multicolumn{1}{c|}{90.44} 
& 0.1498 & \multicolumn{1}{c|}{80.78} 
& 0.4922 & \multicolumn{1}{c}{84.27} \\

\multicolumn{1}{c|}{FinalMLP \cite{10.1609/aaai.v37i4.25577}} 
& 0.3718 & \multicolumn{1}{c|}{79.35} 
& 0.4373 & \multicolumn{1}{c|}{81.45} 
& 0.3058 & \multicolumn{1}{c|}{90.52} 
& 0.1497 & \multicolumn{1}{c|}{80.78} 
& 0.4822 & \multicolumn{1}{c}{85.10} \\

\multicolumn{1}{c|}{FINAL \cite{10.1145/3539618.3591988}} 
& 0.3712 & \multicolumn{1}{c|}{79.41} 
& 0.4371 & \multicolumn{1}{c|}{81.49}
& 0.3035 & \multicolumn{1}{c|}{90.53} 
& 0.1498 & \multicolumn{1}{c|}{80.74} 
& 0.4800 & \multicolumn{1}{c}{85.14} \\

\multicolumn{1}{c|}{RFM \cite{10.1145/3637528.3671740}} 
& 0.3723 & \multicolumn{1}{c|}{79.24} 
& 0.4374 & \multicolumn{1}{c|}{81.47} 
& 0.3048 & \multicolumn{1}{c|}{90.51} 
& 0.1506 & \multicolumn{1}{c|}{80.73} 
& 0.4853 & \multicolumn{1}{c}{84.70} \\

\multicolumn{1}{c|}{QNN \cite{10.1145/3711896.3737106}} 
& 0.3712 & \multicolumn{1}{c|}{79.47} 
& \underline{0.4358} & \multicolumn{1}{c|}{\underline{81.63}} 
& \underline{0.2960} & \multicolumn{1}{c|}{\underline{90.87}} 
& 0.1506 & \multicolumn{1}{c|}{80.82} 
& \underline{0.4730} & \multicolumn{1}{c}{\underline{85.76}} \\

\multicolumn{1}{c|}{FCN \cite{10.1145/3770854.3780177}} 
& \underline{0.3702} & \multicolumn{1}{c|}{\underline{79.66}} 
& 0.4359 & \multicolumn{1}{c|}{81.62} 
& 0.3001 & \multicolumn{1}{c|}{90.74} 
& \underline{0.1494} & \multicolumn{1}{c|}{\underline{80.86}} 
& 0.4746 & \multicolumn{1}{c}{85.72} \\

\hline

\multicolumn{1}{c|}{\textbf{RECAP (ours)}} 
& \textbf{0.3702*} & \multicolumn{1}{c|}{\textbf{79.92*}} 
& \textbf{0.4345*} & \multicolumn{1}{c|}{\textbf{81.76*}} 
& \textbf{0.2951*} & \multicolumn{1}{c|}{\textbf{90.88*}} 
& \textbf{0.1494*} & \multicolumn{1}{c|}{\textbf{81.41*}} 
& \textbf{0.4669*} & \multicolumn{1}{c}{\textbf{86.06*}} \\

\Xhline{1px}
\end{tabular}
}
\vspace{-20pt}
\label{implicit}
\end{table*}

We evaluate RECAP with three questions in mind:
\emph{(1)} does estimator scaling improve predictive performance across standard CTR benchmarks;

\emph{(2)} are these gains explained by the non-shared predictive variation characterized in Section~\ref{sec:estimator_scaling}; and
\emph{(3)} can RECAP realize these gains without the deployment cost of a conventional ensemble?

\textbf{Datasets and metrics: }We evaluate on five public CTR benchmarks: Avazu \cite{10.1145/3459637.3482486}, Criteo \cite{10.1145/3459637.3482486}, MovieLens-1M (ML-1M) \cite{10.1145/3357384.3357925}, KDD12 \cite{10.1145/3357384.3357925}, and KKBox \cite{10.1145/3477495.3531723}. We follow the preprocessing and evaluation protocol defined in the FuxiCTR framework which is used by prior open CTR benchmarks including FCN~\cite{10.1145/3459637.3482486,10.1145/3770854.3780177,10.1145/3477495.3531723} . We report Area Under the ROC Curve (AUC), where higher is better, and binary cross-entropy (Logloss), where lower is better. Because improvements in mature CTR benchmarks are often small in absolute magnitude, we report AUC in percentage points throughout the paper.

\textbf{Baselines: }We compare against representative factorization, deep interaction, and explicit cross-network models, including DNN \cite{10.1145/2959100.2959190}, PNN \cite{7837964}, Wide \& Deep \cite{10.1145/2988450.2988454}, DeepFM \cite{3172077.3172127}, DCN \cite{10.1145/3124749.3124754}, xDeepFM \cite{10.1145/3219819.3220023}, AutoInt \cite{10.1145/3357384.3357925}, AFN \cite{cheng2020adaptivefactorizationnetworklearning}, DCNv2 \cite{10.1145/3442381.3450078}, EDCN \cite{10.1145/3459637.3481915}, MaskNet \cite{wang2021masknetintroducingfeaturewisemultiplication}, EulerNet \cite{10.1145/3539618.3591681}, FinalMLP \cite{10.1609/aaai.v37i4.25577}, FINAL \cite{10.1145/3539618.3591988}, RFM \cite{10.1145/3637528.3671740}, QNN \cite{10.1145/3711896.3737106}, FCN \cite{10.1145/3770854.3780177}. FCN is our primary reference architecture because it provides a strong recent explicit-interaction baseline and directly scales interaction capacity through complementary linear and exponential cross networks.

\vspace{-5pt}
\subsection{RECAP Improves CTR Prediction Across Benchmarks}
\label{sec:main_results}
\vspace{-5pt}
Table~\ref{tab:main_results} compares RECAP with representative CTR prediction models across five public benchmarks. RECAP achieves the highest AUC among the compared methods on all five datasets. Relative to the strongest compared baseline on each dataset, RECAP improves AUC by $0.26$, $0.13$, $0.01$, $0.55$, and $0.30$ points on Avazu, Criteo, ML-1M, KDD12, and KKBox, respectively. RECAP also improves or matches Logloss on all of the datasets as compared to the next best baseline. These results indicate that the benefit of estimator scaling is not restricted to the Criteo diagnostic setting used in our analysis, but transfers across datasets with substantially different sparsity and scale.

\begin{figure*}[t]
    \centering

    \begin{minipage}[t]{0.47\textwidth}
        \vspace{0pt}
        \centering

        \vspace{1.7em}

        \small
        \setlength{\tabcolsep}{3.5pt}
        \renewcommand{\arraystretch}{1.15}

        \begin{tabular}{lcccc}
            \toprule
            \textbf{Model}
            & \textbf{EMA}
            & \textbf{KD}
            & \textbf{Routes}
            & \textbf{AUC} \\
            \midrule

            Recursive backbone
            & -- & -- & --
            & 81.59 \\

             + trajectory scaling
            & \checkmark & -- & --
            & 81.65 \\

            + cross-run scaling
            & -- & \checkmark & --
            & 81.68 \\

            + route scaling
            & -- & -- & \checkmark
            & 81.61 \\

            \textbf{RECAP}
            & \checkmark & \checkmark & \checkmark
            & \textbf{81.76} \\

            \bottomrule
        \end{tabular}

        \vspace{0.5em}

        \footnotesize
        KD denotes ensemble distillation; Routes denotes
        inference-time route aggregation.
        \vspace{1.4em}
        \captionof{table}{\textbf{Ablation of estimator-scaling components on the recursive backbone.} Trajectory scaling through EMA, cross-run scaling through ensemble distillation, and route scaling each improve performance individually, while combining all three yields the highest AUC.}
        \label{tab:recap_ablation}
    \end{minipage}
    \hfill
    \begin{minipage}[t]{0.49\textwidth}
        \vspace{0pt}
        \centering

        \vspace{0.3em}

        \includegraphics[width=0.9\linewidth]{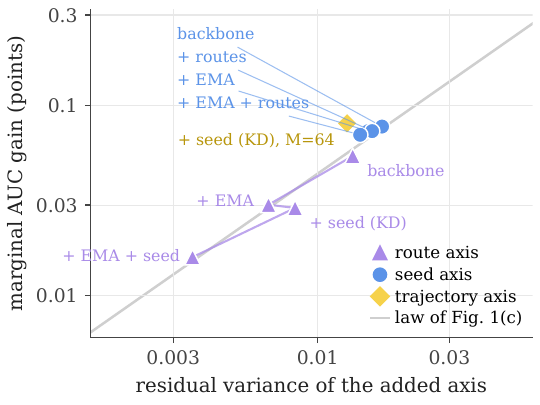}
        \vspace{-10pt}
        \captionof{figure}{\textbf{Marginal estimator-scaling gains track residual predictive variation.} Across trajectory, seed, and inference-time route estimators, the marginal AUC improvement from adding an estimator axis increases with the residual non-shared variance available along that axis.}
        \label{fig:marginal_scaling}
    \end{minipage}
    \vspace{-25pt}
   
\end{figure*}
\vspace{-10pt}
\subsection{Non-Shared Predictive Variation Explains Estimator-Scaling Gains}
\label{sec:variance_results}

Section~\ref{sec:estimator_scaling} makes two related predictions about estimator scaling. First, the standalone benefit of scaling a particular estimator source should depend on the non-shared predictive variation accessible along that axis. Second, when multiple estimator sources are combined, the marginal benefit of adding a new axis should depend on the residual variation that remains after the existing axes have already been incorporated. We test both predictions using the estimator families introduced in Figure~\ref{fig:intro}(b).

\textbf{Standalone estimator scaling: } Figure~\ref{fig:intro}(c) compares the gain from averaging estimators within each axis against the corresponding reducible predictive variation. Despite their different origins, the three estimator families follow a common trend: larger non-shared variation is associated with larger averaging gains. As shown in Figure~\ref{fig:intro}(b), independent training runs occupy the high-variation, high-gain regime, checkpoints from a single optimization trajectory expose less reducible variation and provide smaller gains, and alternative inference-time routes through a shared model expose the least variation and provide the smallest gains. Equation~\ref{eq:curvature_weighted_variance} formally predicts expected BCE improvement. Empirically, AUC exhibits the same ordering: estimator families with greater reducible variation provide larger averaging gains. Results on additional datasets can be found in Appendix \ref{app:diagnosis}.

\textbf{Marginal gains from combining estimator axes: }We next test the marginal improvement prediction from Section~\ref{sec:residual_variance}. Figure~\ref{fig:marginal_scaling} plots the marginal AUC improvement obtained by adding an estimator axis to different existing configurations against the residual non-shared variance available to that axis. The same estimator source can provide different gains depending on what has already been averaged: when previous mechanisms remove overlapping variation, less residual variation remains and the incremental benefit decreases. Across seed, trajectory, and route estimators, configurations with greater residual variation consistently yield larger marginal improvements. This relationship explains why the gains from different estimator-scaling mechanisms are complementary but not simply additive, and is consistent with the decomposition in Eq.~\ref{eq:two_axis_total_variance}. Table \ref{tab:recap_ablation} further confirms complementarity within RECAP: each estimator source improves the same recursive backbone individually, while combining trajectory, cross-run, and route scaling yields the strongest performance.

\begin{wrapfigure}{r}{0.45\textwidth}
\vspace{-10pt}
\begin{minipage}{0.45\textwidth}
\small
        \setlength{\tabcolsep}{3.5pt}
        \renewcommand{\arraystretch}{1.15}
        \resizebox{\linewidth}{!}{
        \begin{tabular}{lccc}
            \toprule
            \textbf{Model}
            & \textbf{EMA}
            & \textbf{KD}
            & \textbf{AUC} \\
            \midrule

            FCN
            & -- & --
            & 81.62 \\

             + trajectory scaling
            & \checkmark & --
            & 81.67 \\

            + cross-run scaling
            & -- & \checkmark
            & 81.69 \\

            + cross-run + trajectory scaling
            & \checkmark & \checkmark
            & 81.74 \\

            \bottomrule
        \end{tabular}
        }
        \vspace{0.5em}

        \captionof{table}{\textbf{Estimator scaling improves the FCN backbone.}
Trajectory scaling through EMA and cross-run scaling through ensemble distillation each improve AUC individually, while combining both yields the strongest performance.} 
\label{tab:fcn_scaling}
\vspace{-20pt}
\end{minipage}
\end{wrapfigure}
\vspace{-5pt}
\subsection{Estimator Scaling Is Not Specific to the Recursive Backbone}
\label{sec:fcn_scaling}
\vspace{-5pt}

To separate estimator-scaling gains from the recursive architecture, we apply the same training schedule, EMA, and ensemble-distillation objective to the FCN backbone. As shown in Table \ref{tab:fcn_scaling}, both trajectory scaling through EMA and cross-run scaling through distillation improve FCN, confirming that estimator scaling is a general principle rather than a RECAP-specific effect. The key contribution of the recursive architecture is to introduce an additional \emph{route-scaling axis}: weight sharing keeps interaction computation parameter-efficient, while lightweight route-specific adaptations define multiple related predictors within a single model. RECAP therefore scales estimators across training trajectories, independent runs, and alternative computation routes, with the latter providing a low-cost source of estimator diversity unavailable to the FCN backbone.

\vspace{-5pt}
\subsection{RECAP Shifts the Performance--Parameter Frontier}

\begin{wrapfigure}{r}{0.45\textwidth}
\vspace{-10pt}
\begin{minipage}{0.45\textwidth}
\includegraphics[width=\linewidth]{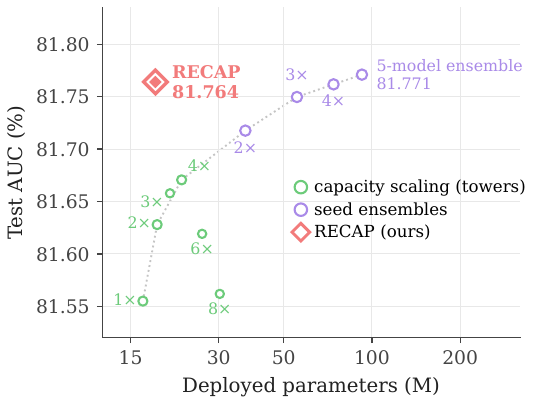}
\captionof{figure}{\textbf{RECAP improves the deployment performance--parameter frontier.} Increasing interaction capacity through additional towers yields diminishing returns, while seed ensembling continues to improve AUC at substantially higher deployment cost. RECAP reaches near five-model ensemble performance while retaining approximately a single-model parameter footprint.}
\label{fig:pareto}
\vspace{-10pt}
\end{minipage}
\end{wrapfigure}

The performance--parameter analyses in Figures~\ref{fig:intro}(d) and~\ref{fig:pareto} provide complementary views of RECAP's deployment efficiency. Against a broad set of CTR architectures, RECAP substantially outperforms single-model baselines at a comparable parameter budget and approaches the five-model FCN ensemble ($81.764$ versus $81.771$ AUC) while using only a fraction of its deployed parameters. The controlled scaling trajectories in Figure~\ref{fig:pareto} further show that increasing interaction capacity through additional towers quickly yields diminishing returns, whereas independent-model ensembling continues to improve performance but at roughly linear deployment cost. RECAP lies above the interaction-capacity scaling trajectory and reaches the performance regime of substantially larger seed ensembles at approximately a single-model parameter budget. Together, these results suggest that once interaction-capacity scaling becomes inefficient, allocating resources toward estimator scaling provides a more favorable accuracy--parameter trade-off. RECAP captures much of this estimator-scaling benefit while avoiding the deployment cost of maintaining multiple full models.

\vspace{-5pt}
\section{Conclusion}
In this work, we tackle the problem of CTR prediction and show that scaling interaction capacity in CTR models can exhibit diminishing predictive returns even when estimator diversity remains exploitable. This motivates \emph{estimator scaling} as a complementary axis, with gains governed by the non-shared predictive variation available to aggregation. Building on this view, RECAP combines cross-run distillation, trajectory averaging, and route-based estimator scaling within a parameter-efficient recursive backbone. Across five CTR benchmarks, RECAP achieves state-of-the-art performance, while approaching five-model ensemble performance at approximately a single-model parameter budget.

\bibliography{iclr2027_conference}
\bibliographystyle{iclr2027_conference}

\newpage 

\appendix

\section{Diagnosis Experiments on Additional Datasets}
\label{app:diagnosis}

\begin{figure}[H]
    \centering
    \includegraphics[width=\linewidth]{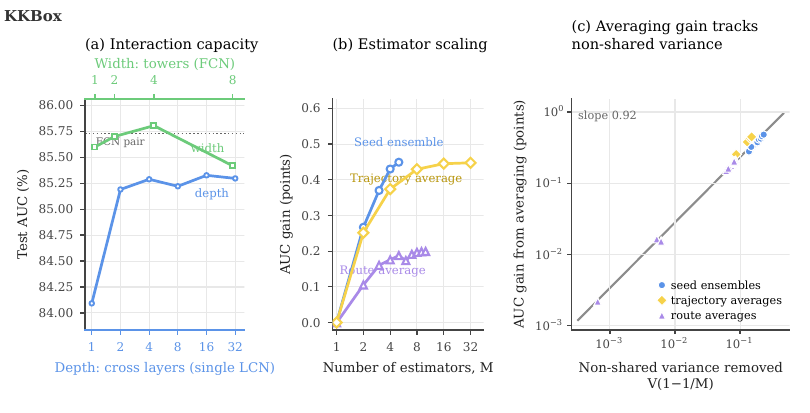}
    \caption{
    \textbf{Diagnostic analysis of interaction-capacity and estimator scaling on KKBox.}
    The same qualitative picture holds on KKBox: capacity scaling saturates, estimator scaling continues to help, and the gains from averaging track the amount of non-shared predictive variation exposed by each estimator source.
    }
    \label{fig:diag_kkbox}
\end{figure}

\begin{figure}[H]
    \centering
    \includegraphics[width=\linewidth]{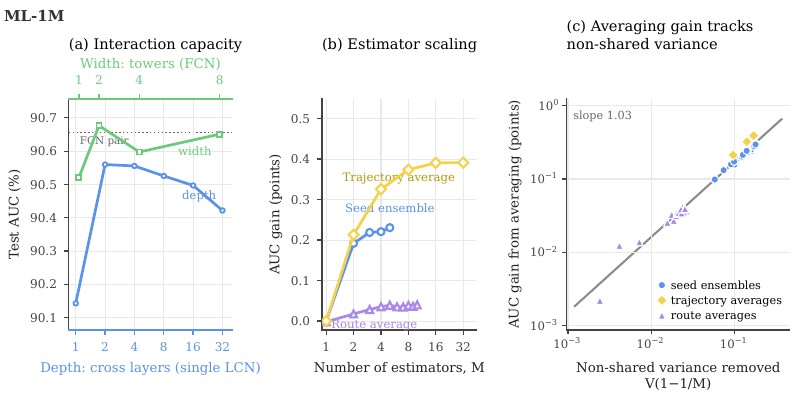}
    \caption{
    \textbf{Diagnostic analysis of interaction-capacity and estimator scaling on ML-1M.}
    ML-1M also exhibits diminishing returns from additional interaction capacity together with continued gains from estimator scaling. The observed averaging improvements remain strongly aligned with the amount of non-shared predictive variation available across estimator families.
    }
    \label{fig:diag_ml1m}
\end{figure}

\begin{figure}[H]
    \centering
    \includegraphics[width=\linewidth]{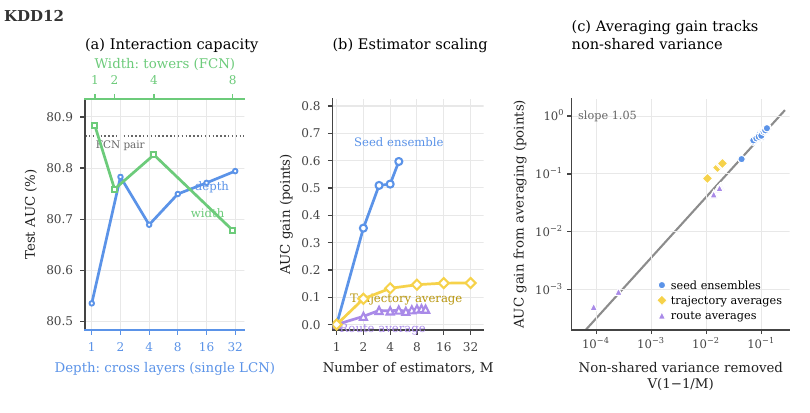}
    \caption{
    \textbf{Diagnostic analysis of interaction-capacity and estimator scaling on KDD12.}
    Increasing interaction capacity exhibits diminishing returns, whereas estimator scaling continues to provide gains. As in the main text, the improvement from averaging is closely associated with the non-shared predictive variation available to aggregation.
    }
    \label{fig:diag_kdd}
\end{figure}

\begin{figure}[H]
    \centering
    \includegraphics[width=\linewidth]{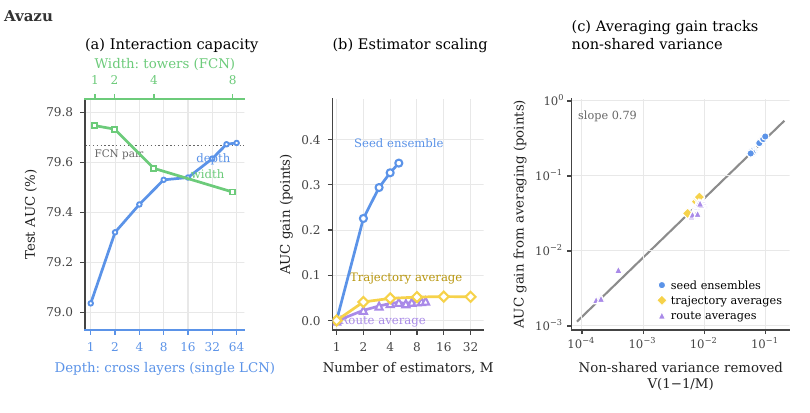}
    \caption{
    \textbf{Diagnostic analysis of interaction-capacity and estimator scaling on Avazu.}
    Increasing interaction capacity exhibits diminishing returns, whereas estimator scaling continues to provide gains. As in the main text, the improvement from averaging is closely associated with the non-shared predictive variation available to aggregation.
    }
    \label{fig:diag_avazu}
\end{figure}

To test whether the diagnostic findings from Criteo generalize beyond a single benchmark, we repeat the same analysis on additional CTR datasets. Figures~\ref{fig:diag_kkbox}, \ref{fig:diag_ml1m}, \ref{fig:diag_kdd} and \ref{fig:diag_avazu} report the corresponding results on KKBox, ML-1M, KDD12, and Avazu. As in the main text, these diagnostics compare interaction-capacity scaling against estimator scaling and relate averaging gains to the amount of non-shared predictive variation available across estimator sources.

Across all four datasets, we observe the same qualitative pattern as on Criteo. First, increasing interaction capacity through greater depth or breadth yields diminishing returns once the underlying interaction model becomes sufficiently expressive. Second, averaging additional estimators continues to improve predictive performance after this capacity-scaling regime begins to saturate. Third, the magnitude of these gains is consistently associated with the amount of non-shared predictive variation exposed by each estimator source. These results support the main claim of the paper: interaction-capacity scaling and estimator scaling are complementary, and the usefulness of estimator scaling is governed by the reducible variation available to aggregation.

At the same time, the relative strength of different estimator sources varies across datasets. On some datasets, independent training runs provide the largest gains, while on others trajectory-based averaging is comparatively stronger. This variation is consistent with the theory in Section~\ref{sec:estimator_scaling}: the benefit of an estimator axis is determined not by its label (seed, trajectory, or route) but by how much non-shared predictive variation it contributes. This further motivates RECAP's use of multiple estimator axes rather than relying on a single source of diversity.

Taken together, these additional results show that the diagnosis in Figure~\ref{fig:intro} is not specific to Criteo. Across diverse CTR benchmarks, increasing the capacity of a single predictor eventually becomes inefficient, while estimator scaling remains beneficial whenever additional non-shared predictive variation is available. This consistency across datasets provides further support for the estimator-scaling perspective developed in the main paper.

\section{Additional Theory for Estimator Scaling}
\label{app:variance_theory}

This appendix provides the derivations underlying Section~\ref{sec:estimator_scaling}. We first extend the equicorrelated analysis in the main text to arbitrary estimator covariance, and then derive the local relationship between reducible predictive variation and expected binary cross-entropy. Throughout this appendix, estimator variation is defined conditionally on the input $x$ and for a fixed interaction capacity $c$.

\subsection{General Covariance Formulation}
\label{app:general_covariance}

For a fixed input $x$ and interaction capacity $c$, let
\begin{equation}
    \mathbf{Z}_c(x)
    =
    [Z_{c,1}(x),\ldots,Z_{c,M}(x)]^\top
\end{equation}
denote the logits of $M$ estimators, with conditional covariance matrix
\begin{equation}
    \Sigma_c(x)
    =
    \operatorname{Cov}\!\left(
        \mathbf{Z}_c(x)\mid x
    \right).
\end{equation}
Their mean-logit aggregate is
\begin{equation}
    \bar Z_{c,M}(x)
    =
    \frac{1}{M}
    \mathbf{1}^\top
    \mathbf{Z}_c(x),
\end{equation}
and therefore
\begin{equation}
    \operatorname{Var}
    \left(
        \bar Z_{c,M}(x)\mid x
    \right)
    =
    \frac{1}{M^2}
    \mathbf{1}^\top
    \Sigma_c(x)
    \mathbf{1}.
    \label{eq:app_general_mean_variance}
\end{equation}

The average conditional variance of an individual estimator is
\begin{equation}
    V_{\mathrm{ind}}(c;x)
    =
    \frac{1}{M}
    \operatorname{tr}\!\left(
        \Sigma_c(x)
    \right).
\end{equation}
We define the variation removed by averaging as
\begin{equation}
    V_{\mathrm{ns}}(c,M;x)
    =
    \frac{1}{M}
    \operatorname{tr}\!\left(
        \Sigma_c(x)
    \right)
    -
    \frac{1}{M^2}
    \mathbf{1}^\top
    \Sigma_c(x)
    \mathbf{1}.
    \label{eq:app_nonshared_variance}
\end{equation}

Equation~\ref{eq:app_nonshared_variance} shows that averaging gains depend jointly on estimator variance and covariance. A collection of estimators can exhibit substantial individual variation while providing little reducible variation if that variation is strongly shared.

For the equicorrelated setting used in the main text, suppose
\begin{equation}
    \operatorname{Var}
    \left(
        Z_{c,m}(x)\mid x
    \right)
    =
    \tau_c^2(x),
    \qquad
    \operatorname{Corr}
    \left(
        Z_{c,i}(x),Z_{c,j}(x)\mid x
    \right)
    =
    \rho_c(x),
    \quad i\neq j.
\end{equation}
Positive semidefiniteness requires
\begin{equation}
    \rho_c(x)
    \geq
    -\frac{1}{M-1}.
\end{equation}
Substituting into Eq.~\ref{eq:app_general_mean_variance} gives
\begin{equation}
    \operatorname{Var}
    \left(
        \bar Z_{c,M}(x)\mid x
    \right)
    =
    \tau_c^2(x)
    \left(
        \rho_c(x)
        +
        \frac{1-\rho_c(x)}{M}
    \right).
\end{equation}
Since a single estimator has variance $\tau_c^2(x)$, the variation removed by averaging is
\begin{equation}
    V_{\mathrm{ns}}(c,M;x)
    =
    \tau_c^2(x)
    \left(
        1-\rho_c(x)
    \right)
    \left(
        1-\frac{1}{M}
    \right),
\end{equation}
recovering Eq.~\ref{eq:variance_removed}.

\subsection{From Reducible Variation to Binary Cross-Entropy}
\label{app:variance_to_loss}

We next derive Eqs.~\ref{eq:capacity_estimator_decomposition} and~\ref{eq:curvature_weighted_variance}. Recall that binary cross-entropy in logit space is
\begin{equation}
    \ell(y,z)
    =
    \log(1+\exp(z))-yz.
\end{equation}
For a fixed $x$, write the $M$-estimator aggregate as
\begin{equation}
    \bar Z_{c,M}(x)
    =
    \mu_c(x)
    +
    \epsilon_{c,M}(x),
    \qquad
    \mathbb{E}
    \left[
        \epsilon_{c,M}(x)\mid x
    \right]
    =
    0,
\end{equation}
where $\mu_c(x)=\mathbb{E}[Z_c(x)\mid x]$ is the estimator-family mean.

A second-order Taylor expansion around $\mu_c(x)$ gives
\begin{align}
    \ell
    \left(
        y,\mu_c+\epsilon_{c,M}
    \right)
    ={}&
    \ell(y,\mu_c)
    +
    \ell'(y,\mu_c)\epsilon_{c,M}
    \nonumber\\
    &+
    \frac{1}{2}
    \ell''(y,\mu_c)
    \epsilon_{c,M}^2
    +
    O
    \left(
        |\epsilon_{c,M}|^3
    \right).
    \label{eq:app_taylor_loss}
\end{align}

Taking expectation over estimator randomness conditional on $x$ eliminates the first-order term:
\begin{align}
    \mathbb{E}
    \left[
        \ell
        \left(
            y,\bar Z_{c,M}(x)
        \right)
        \mid x
    \right]
    \approx{}&
    \ell
    \left(
        y,\mu_c(x)
    \right)
    \nonumber\\
    &+
    \frac{1}{2}
    \ell''
    \left(
        y,\mu_c(x)
    \right)
    \operatorname{Var}
    \left(
        \bar Z_{c,M}(x)\mid x
    \right).
\end{align}

For binary cross-entropy,
\begin{equation}
    \ell''(y,z)
    =
    \sigma(z)
    \left(
        1-\sigma(z)
    \right),
\end{equation}
which is independent of the label $y$. Averaging over the data distribution therefore gives
\begin{equation}
    \mathcal{L}(c,M)
    \approx
    \mathcal{L}(\mu_c)
    +
    \frac{1}{2}
    \mathbb{E}_{x}
    \left[
        \sigma(\mu_c(x))
        \left(
            1-\sigma(\mu_c(x))
        \right)
        \operatorname{Var}
        \left(
            \bar Z_{c,M}(x)\mid x
        \right)
    \right],
    \label{eq:app_capacity_estimator_decomposition}
\end{equation}
which is Eq.~\ref{eq:capacity_estimator_decomposition} in the main text, up to the omitted third-order remainder.

Comparing a single estimator with an $M$-estimator aggregate and using Eq.~\ref{eq:variance_removed}, the corresponding reduction in expected BCE is
\begin{equation}
    \Delta\mathcal{L}(c,M)
    \approx
    \frac{1}{2}
    \mathbb{E}_{x}
    \left[
        \sigma(\mu_c(x))
        \left(
            1-\sigma(\mu_c(x))
        \right)
        V_{\mathrm{ns}}(c,M;x)
    \right].
    \label{eq:app_curvature_weighted_variance}
\end{equation}

Thus, BCE improvement is locally proportional to \emph{curvature-weighted reducible predictive variation}. When the estimator families being compared have similar mean predictions, the curvature term $\sigma(\mu_c(x))(1-\sigma(\mu_c(x)))$ is approximately shared, so differences in averaging benefit are primarily governed by $V_{\mathrm{ns}}(c,M;x)$. This motivates the reducible-variance diagnostic used in the main experiments. The theoretical proportionality applies directly to expected BCE; the corresponding association with AUC reported in the main text is empirical rather than implied by this derivation.

The approximation is most accurate when estimator disagreement around $\mu_c(x)$ is modest. For broader or heavy-tailed estimator distributions, the omitted third-order and higher-order terms may become non-negligible.

\subsection{Exact Identity for Literal Mean-Logit Aggregation}
\label{app:exact_logit_special_case}

For literal mean-logit aggregation, binary cross-entropy also admits an exact inequality. For a fixed example $(x,y)$ and estimator logits $z_1,\ldots,z_M$, let
\begin{equation}
    \bar z
    =
    \frac{1}{M}
    \sum_{m=1}^{M}z_m.
\end{equation}
Since
\begin{equation}
    \ell(y,z)
    =
    \operatorname{softplus}(z)-yz,
\end{equation}
we obtain
\begin{align}
    \frac{1}{M}
    \sum_{m=1}^{M}
    \ell(y,z_m)
    -
    \ell(y,\bar z)
    &=
    \frac{1}{M}
    \sum_{m=1}^{M}
    \operatorname{softplus}(z_m)
    -
    \operatorname{softplus}(\bar z)
    \nonumber\\
    &\geq 0,
    \label{eq:app_exact_logit_gap}
\end{align}
where the inequality follows from convexity of $\operatorname{softplus}$.

Equation~\ref{eq:app_exact_logit_gap} shows that literal mean-logit aggregation cannot have higher BCE than the average BCE of its constituent logits on the same example. This exact result does not, however, quantify the gain in terms of reducible variance, nor does it directly characterize parameter averaging such as EMA or a distilled student that only approximates an ensemble prediction. For this reason, the main text uses the local variance formulation as the common explanatory framework across estimator sources.

\section{Combining Multiple Estimator Axes}
\label{app:multi_axis_variance}

\subsection{Two-Axis Decomposition}
\label{app:two_axis_variance}

Let $A$ and $B$ denote two estimator sources and let $Z_{A,B}(x)$ denote the corresponding random logit for a fixed input $x$. Applying the law of total variance while treating $A$ as the first averaging axis gives
\begin{equation}
    \operatorname{Var}
    \left(
        Z_{A,B}(x)\mid x
    \right)
    =
    \mathbb{E}_{B}
    \left[
        \operatorname{Var}_{A}
        \left(
            Z_{A,B}(x)\mid B,x
        \right)
    \right]
    +
    \operatorname{Var}_{B}
    \left[
        \mathbb{E}_{A}
        \left(
            Z_{A,B}(x)\mid B,x
        \right)
    \right].
    \label{eq:app_two_axis_variance}
\end{equation}

Define
\begin{equation}
    V_A(x)
    =
    \mathbb{E}_{B}
    \left[
        \operatorname{Var}_{A}
        \left(
            Z_{A,B}(x)\mid B,x
        \right)
    \right],
\end{equation}
which is the variation directly accessible to averaging along axis $A$. After averaging over $A$, the remaining predictor is
\begin{equation}
    \mu_A(B;x)
    =
    \mathbb{E}_{A}
    \left[
        Z_{A,B}(x)\mid B,x
    \right],
\end{equation}
whose residual variation across $B$ is
\begin{equation}
    V_{B\mid A}(x)
    =
    \operatorname{Var}_{B}
    \left[
        \mu_A(B;x)
    \right].
\end{equation}
Hence
\begin{equation}
    \operatorname{Var}
    \left(
        Z_{A,B}(x)\mid x
    \right)
    =
    V_A(x)+V_{B\mid A}(x).
\end{equation}

Up to the same local BCE-curvature weighting derived in Appendix~\ref{app:variance_to_loss}, $V_A(x)$ determines the standalone loss reduction available from averaging along axis $A$, while $V_{B\mid A}(x)$ determines the residual variation available to a subsequent estimator axis $B$. This is the basis for the marginal-gain analysis in Section~\ref{sec:variance_results}.

The decomposition is order-dependent. Reversing the averaging order gives
\begin{equation}
    \operatorname{Var}
    \left(
        Z_{A,B}(x)\mid x
    \right)
    =
    V_B(x)+V_{A\mid B}(x),
\end{equation}
and in general
\begin{equation}
    V_A(x)\neq V_{A\mid B}(x),
    \qquad
    V_B(x)\neq V_{B\mid A}(x).
\end{equation}
The total predictive variance is unchanged, but its attribution to the estimator axes depends on the averaging order.

\subsection{Three-Axis Decomposition for RECAP}
\label{app:three_axis_variance}

Let $S$, $T$, and $R$ denote independent training runs, checkpoints along a training trajectory, and inference-time routes, respectively. For a fixed input $x$, let $Z_{S,T,R}(x)$ denote the resulting random logit. We use the operational nesting order
\begin{equation}
    R
    \rightarrow
    T
    \rightarrow
    S,
\end{equation}
corresponding to route averaging first, trajectory averaging next, and cross-run averaging last.

Applying the law of total variance recursively gives
\begin{align}
    \operatorname{Var}
    \left(
        Z_{S,T,R}(x)\mid x
    \right)
    ={}&
    \underbrace{
        \mathbb{E}_{S,T}
        \left[
            \operatorname{Var}_{R}
            \left(
                Z_{S,T,R}(x)\mid S,T,x
            \right)
        \right]
    }_{V_R(x)}
    \nonumber\\
    &+
    \underbrace{
        \mathbb{E}_{S}
        \left[
            \operatorname{Var}_{T}
            \left(
                \mathbb{E}_{R}
                \left[
                    Z_{S,T,R}(x)\mid S,T,x
                \right]
                \mid S,x
            \right)
        \right]
    }_{V_{T\mid R}(x)}
    \nonumber\\
    &+
    \underbrace{
        \operatorname{Var}_{S}
        \left[
            \mathbb{E}_{T,R}
            \left(
                Z_{S,T,R}(x)\mid S,x
            \right)
        \right]
    }_{V_{S\mid T,R}(x)}.
    \label{eq:app_three_axis_decomposition}
\end{align}

The three terms have direct operational interpretations:
\begin{itemize}
    \item $V_R(x)$ is variation accessible to inference-time route averaging;
    \item $V_{T\mid R}(x)$ is trajectory variation remaining after route averaging;
    \item $V_{S\mid T,R}(x)$ is cross-run variation remaining after both route and trajectory averaging.
\end{itemize}

Equation~\ref{eq:app_three_axis_decomposition} explains why gains from different estimator sources need not add linearly. If two axes expose overlapping predictive variation, averaging one reduces the residual variation available to the next; if they expose largely distinct variation, their gains can be closer to additive. There are $3! = 6$ possible nestings of $(S,T,R)$. Their component values generally differ, although all decompositions sum to the same
total predictive variance. The decomposition is an output-space attribution of the variation associated with the three estimator sources; it should not be interpreted as the literal sequence of operations implemented by RECAP. For a fixed attribution convention, we use the nesting order \(R\to T\to S\), attributing route-accessible variation first, followed by residual trajectory and cross-run variation.

\section{Hyperparameter Sensitivity}
\label{app:hyperparams}

\begin{table}[H]
\centering
\caption{
Hyperparameter sensitivity of RECAP on Criteo.
We vary one method-specific hyperparameter at a time while keeping all other settings fixed. Logloss is the test logloss after validation-fitted affine calibration.
}
\label{tab:hyperparameter_sensitivity}

\begin{tabular}{llcc}
\toprule
Hyperparameter & Value & AUC (\%) $\uparrow$ & Logloss $\downarrow$ \\
\midrule

\multirow{5}{*}{EMA decay $\beta$}
& 0.9900 & 81.735 & 0.43487 \\
& 0.9950 & 81.744 & 0.43482 \\
& 0.9990 & 81.760 & 0.43461 \\
& \textbf{0.9995} & \textbf{81.765} & \textbf{0.43456} \\
& 0.9999 & 81.770 & 0.43450 \\
\midrule

\multirow{5}{*}{KD weight $\lambda_{\mathrm{KD}}$}
& 0.00 & 81.665 & 0.43535 \\
& 0.25 & 81.743 & 0.43461 \\
& 0.50 & 81.766 & 0.43443 \\
& \textbf{1.00} & \textbf{81.765} & \textbf{0.43456} \\
& 2.00 & 81.763 & 0.43454 \\
\midrule

\multirow{5}{*}{Teacher seeds $M_{\mathrm{seed}}$}
& 1 & 81.704 & 0.43502 \\
& 2 & 81.744 & 0.43478 \\
& 3 & 81.756 & 0.43469 \\
& \textbf{5} & \textbf{81.765} & \textbf{0.43456} \\
& 8 & 81.768 & 0.43454 \\
\midrule

\multirow{4}{*}{LoRA rank $r_{\mathrm{LoRA}}$}
& 24 & 81.763 & 0.43459 \\
& 48 & 81.765 & 0.43444 \\
& \textbf{96} & \textbf{81.765} & \textbf{0.43456} \\
& 192 & 81.767 & 0.43444 \\
\midrule

\multirow{4}{*}{Route adapters $M_{\mathrm{route}}$}
& 2 & 81.759 & 0.43452 \\
& 4 & 81.767 & 0.43448 \\
& \textbf{8} & \textbf{81.765} & \textbf{0.43456} \\
& 16 & 81.764 & 0.43444 \\
\bottomrule

\end{tabular}
\end{table}

We study the sensitivity of RECAP to its main method-specific hyperparameters on Criteo. We vary one hyperparameter at a time while keeping all remaining settings fixed to the default configuration used in the main experiments. We consider five quantities: the EMA decay $\beta$, the distillation weight $\lambda_{\mathrm{KD}}$, the number of teacher seed models $M_{\mathrm{seed}}$, the LoRA rank $r_{\mathrm{LoRA}}$, and the number of route adapters $M_{\mathrm{route}}$. These parameters control trajectory averaging, the strength and breadth of cross-run distillation, and the capacity of the route-scaling mechanism. Unless otherwise specified, each configuration is evaluated using the same training protocol and data split as the main experiments.

Table~\ref{tab:hyperparameter_sensitivity} reports the resulting test AUC and Logloss. Overall, RECAP is stable across a broad range of settings, with the default configuration lying in regions of consistently strong performance rather than at isolated optima. Performance varies only modestly across EMA decay values, while nonzero distillation weights consistently outperform the no-distillation setting. Increasing the number of teacher seeds improves performance with diminishing returns, consistent with the saturation behavior observed for independent estimators. Likewise, increasing the LoRA rank or the number of route adapters beyond moderate values provides little additional improvement.

These results indicate that RECAP does not depend on narrowly tuned hyperparameters. In particular, its performance remains robust across the trajectory-averaging, cross-run distillation, and route-capacity settings considered here.

\end{document}